# A Survey on Why-Type Question Answering Systems

Manvi Breja and Sanjay Kumar Jain

(National Institute of Technology Kurukshetra, Haryana, India)

Search engines such as Google, Yahoo and Baidu yield information in the form of a relevant set of web pages according to the need of the user. Question Answering Systems reduce the time taken to get an answer, to a query asked in natural language by providing the 'one' most relevant answer. To the best of our knowledge, major research in Why-type questions began in early 2000's and our work on Why-type questions can help explore newer avenues for fact-finding and analysis. The paper presents a survey on Why-type Question Answering System, details the architecture, the processes involved in the system and suggests further areas of research.

1. **Introduction**

Information retrieval is the process of retrieving useful information from documentation/web pages to fulfill users' knowledge demands [Croft and Metzler;2010]. While information retrieval is a broader domain, question answering system (QAS) is a branch that employs the procedures of information retrieval (IR) and natural language processing (NLP) to answer the user's input questions in natural language [A A Stupina; 2016]. Question answering (QA) delivers the exact information in a few sentences, instead of overloading the user with a set of webpages which naturally requires a user's intervention to review. The research on QAS has advanced over the past few decades with the pressing need for more precise answers for the user query.

Questions are classified into various types, namely ones initiating with who, when, what, where, how, and why. Questions with what, when, who, and where are characterized as *factoid questions*, questions with why and how are placed under the umbrella of *non-factoid questions*. Factoid questions are relatively easier to process and are responded to/ answered in a single sentence.

What-type questions seek details of the subject in question; when-type questions are mainly addressed to some time information in the past, present or future; who-type questions are aimed at extracting information on a subject/person/entity; and where-type questions are intended to know the locus of the subject in the question. Kolomiyets; 2011 used named entity tagging to study expected answers of factoid questions. Some of the examples of existing factoid QAS are 'Webclopedia' described by research in (Hovy et. al (2000), 'Mulder' by (Kwok et.al. 2001), 'START' by (Katz 2002), 'Answerbus' discussed in (Zheng 2002) and 'Naluri' in (Wong 2004). Compared to factoid questions, non-factoid questions are more complex and their responses need detailed information and in-depth reasoning about the subject in question. Responses are subjective and may range from a sentence to a document. These questions require advanced NLP techniques such as Pragmatic and discourse analysis as discussed in (Verberne et. al., 2007, 2008, 2010), textual entailment by (Harabagiu and Hickl; 2006, Dagan Ido et. al.; 2015) and lexical semantic modelling in (Daniel Fried et. al. 2015, Jansen etc. al. 2014) to get answered.

In the case of non-factoid questions, relatively less precise outcomes have been observed, mainly because responses vary based on reference and temporal boundaries set by the user in the question. The table below enlists performance of some of the researches done on different modules of QAS. Moreover, because non-factoid questions are subjective, there is the possibility of having numerous answers. Thus the need to develop Why QAS stems from the requirement to have more accurate 'one' answer to queries posed to the system.

| Research authors and year | Module of QAS | Performance of research |
|---|---|---|
| Girju, 2006 | Question classification | Tested on 50 questions with 61% precision |
| Suzan verberne, 2006 | Question classification | 62.2% of questions assigned an answer type correctly |
| Suzan verberne, 2007 | Passage retrieval | 21.5% of the questions, no answer was retrieved in the top-150 results |
| Masaki Murata et. al. , 2008 | Passage Retrieval | Accuracy rate 0.77 |
| Fukumoto, 2007 | Answer candidate extraction | 50% accuracy compared to human ranking |
| Mori et. al. , 2008 | Answer candidate extraction | MRR of 0.307 with no. of well answered questions are 16 out of 33 |
| Reyes et. al, 2008 | Answer candidate extraction | Correct answers at first position: 34%<br>Correct answers at second position: 19%<br>Correct answers at third position: 44%<br>*MRR* is 0.601 |
| Yih et. al; 2013 | Answer candidate extraction | 0.7648 MAP 0.8255 MRR |
| Liu Yang et. al., 2016 | Answer candidate extraction | P@10 – 0.817, MRR-0.4070: (MK+Semantics+Context) with Coord. Ascent, P@10-0.2024, MRR-0.4512: (MK+Semantics+Context) with MART, P@10-0.1939, MRR-0.4030: |

| | | (MK+Semantics+Context) with LambdaMART |
|---|---|---|
| Jong-Hoon Oh et. al.;2016 | Answer candidate extraction | 50.0 P@1 48.9 MAP 75.8 R(P@1) 74.1 R(MAP) 68.0 P@3 75.0 P@5 |
| Jong-Hoon Oh et. al.; 2019 | Answer candidate extraction | 54.8% P@1 52.4% MAP |
| Suzan verberne, 2007 | Answer candidate ranker | success@150 is 78.5% |
| Higashinaka and Isozaki, 2008 | Answer candidate ranker | Mean Reciprocal Rank (top-5) of 0.305, making it presumably the best-performing fully implemented why-QA system. |
| Jong-Hoon Oh et. al., 2012 | Answer Ranking | 64.8% in P@1 and 66.6% in MAP |
| Jong-Hoon Oh et.al; 2013 | Answer candidate ranker | improved P@1 by 4.4%, P@1 41.8, MAP 41.0, 83.2% precision |
| Jansen and Mihai Surdeanu, 2014 | Answer candidate ranker | Performance improvement up to 24% |

*Table 1: Researches in Why-QA enlisting performance*

Several research papers have been instrumental in contributing to the development of non-factoid QAS. This paper surveys the development of Why-QASs, unfolding various important terms that are encountered and illustrating the classification of QAS with its generalized architecture. It is furthered by discussing the different techniques utilized in implementing various modules of Why-type QASs. The paper also describes the metrics used to measure the performance of QAS, and concludes with the subsequent work.

## 2. Digital Assistants vs Question Answering Systems

Google Assistant, Alexa, Siri, and Cortana are some of the common examples of digital assistants prevalent in market today. They can answer any question or execute any command given to it. They are designed to offer functionalities varying with playing songs/video/movie, telling the current weather or predicting weather forecast, setting alarms, making and taking phone calls to even searching the Web for answering trivial questions. They are not only limited to dealing with textual data rather they can deal with audio, images and videos as well. Question Answering system is a broad domain whose one of the famous applications is digital assistant. They utilize both Natural language processing and machine learning to understand the question in natural language and parse it to recognize interesting meaningful patterns using machine learning algorithm. There are two phases of processing information (1) Training (Pre-processing) stage and (2) Process and Decision making stage. Training stage focuses on recognizing Named Entity and Intent from the query given to the system. For example, in the command "Alexa, please have a tea", "tea" is an "Entity" (name, date, location, property) and "have" is a "Intent" (action invoked by the user). Second stage of processing and decision making comprises many text processing stages like Stemming and Lemmatizing, TF-IDF, Co-reference resolution, Part-of-speech (POS) Tagging, Dependency Parsing, Named Entity/Intent Recognition & many more.

They can also address the domain-specific question using the knowledge extracted from the knowledge graph. User question is decomposed into a vector space using semantic similarity matching. Semantics find the meaning and interpretation of the words contained in question. One of the techniques used in semantic analysis is Word sense disambiguation which assigns the meaning to each word based on the context. One of the challenges faced by digital assistants is context and personalization i.e. answering query on the basis of the device from which user is asking, location, and the time of asking etc. Besides the existing challenges, the paper is trying to get into the inner working of question answering system that can answer only Why-type questions accurately.

3. **Important Definitions**

There are some important terminologies need to be understood to get the insights of Why-Question answering system. These contain some of the common terms as well as some research techniques addressed by authors. The brief of the terminologies have been discussed below:

**Natural language processing:** Natural language is a way of communication among humans, either with speech or via text. Natural language processing [Collobert; 2011] aims to bridge the gap between computers and humans, and helps the computer understand, manipulate, and process human knowledge.

**Causality:** It forms the bedrock for answering Why-type questions, which connects one phrase (the cause) with another phrase (the effect). For example, Ravi was suffering from fever, is the cause of his not performing well in exams which is its effect. [Higashinaka and Isozaki; 2008]

**Machine learning:** Machine learning [Mitchell 2013] is a computer science discipline in artificial intelligence, which aims to make computers learn with input data and act in a relevant scenario. It analyses the data, identifies patterns and lastly retrieves decisions without human intervention. It is widely applicable in areas such as recommendation systems, predictive analyser, business intelligence, self-driving cars, and assistive technology.

**Classification:** Classification is a data analysis approach that categorizes numerical data. It assigns a label to a dataset, by determining its features [CC Aggarwal; 2012]. There are many classification algorithms that build the classifier from the training set consisting of tuples with their assigned class labels. The classifier learns from the training data and assigns a class to test data by applying classification rules formulated by the given classification algorithms.

**Clustering:** Clustering is a form of unsupervised learning that identifies similar traits in data points and groups them in a cluster [CC Aggarwal; 2012]. Cluster is an aggregation of items that have similarity among them but dissimilar to the items in other clusters.

**Neural networks:** Artificial Neural Network [PM Buscema 2018] stimulates the process of processing information by a biological neural system. It is a collection of interconnected numerous neurones which learns itself to solve complex problems. Connections between neurons are used to transmit signal from one neuron to another. It understands imprecise data and uses that meaning to recognize patterns and trends. It has various characteristics such as adaptive learning, self-organisation, real time operation, and fault-tolerance.

**Hypothesis:** A hypothesis is used to understand the relationship between variables and is a prediction test for some event/phenomenon. It has to be measurable and clearly understandable. It can be proven to be both right and wrong.

**Opinionated:** Opinionated means having strong opinions. Unlike having an opinion, opinionated has an obstructive aura, which means sticking to one's opinions without considering others' views, opinions and situations.

**Data mining:** Data mining intends to extract useful information from raw data [DT Larose 2014]. It discovers patterns, and identifies relationships and correlations from the massive collection of data sets to solve problems and predict future outcomes.

**Corpus vs Data Set:** Corpus and Data Set both are the terms used for the collection of data. Corpus is a sample that has a wide context and contains general purpose data, whereas a dataset is a sample having a restricted context and refers to some research question.

**Annotation:** Annotation is a process of attaching an explanation to the data. It focuses on understanding text and making notes, to enhance the reader's reaction to the text. It is used to focus the key areas, the main idea and thoughts of the reader.

**Terminologies focused on types of analysis:**

**Lexical analysis:** Lexical analysis [Robert; 2000] is the first component that divides a text into tokens to analyse the structure of the sentence.

**Syntactic analysis:** This is used to determine the grammatical structure of the sentence by analysing the ordering of the words to determine the relationship among them [Robert; 2000]. For example, the sentence "the boy coffee likes" is rejected by the English syntactic analyzer, because the ordering of words is grammatically incorrect.

**Semantic analysis:** This analysis is used to infer the exact meaning of the sentence [Robert; 2000]. It assigns dictionary meaning to the structures returned by the syntactic analyser. For example, "cold tea" is disregarded by the semantic analyzer because there is no dictionary meaning attached with "cold tea".

**Pragmatic analysis:** It is one of the major components of Natural Language processing [WG Lehnert and MH Ringle; 2014]. It tries to understand sentences in different contexts, which requires world knowledge for correct interpretation. It describes how the usage of the sentence in different situations affects their interpretation. For example, the sentence 'close the window' can be taken as a request or a command.

**Discourse analysis:** It is also a component of Natural Language processing which takes the contextual sense [Johnstone 2018]. It describes how the meaning of a sentence is affected by the preceding and succeeding sentences. For example, the sentence "He beats him", prior discourse context will be required to interpret to whom he and him are referring.

**Terminologies focused on techniques used in different modules**

**Named entity tagging:** Named entity tagging [Dozier 2010] is a process that identifies and labels named entities in a text (paragraph or document). Examples of named entities are persons, locality, organizations, timestamp, money, etc. It plays an important role in various fields such as content recommendations, customer feedback, machine learning, etc.

**Textual entailment:** Textual entailment is used to determine the directional link between two text fragments, where the entailing fragment refers to 'text (x)' and the entailed fragment refers to 'hypothesis (y)' [DZ Korman 2018]. It has its applications in various fields such as question answering, document summarization, prediction system, information extraction and machine translation, etc.

**Semantic role labelling:** Semantic role labelling is a process used for semantic representation to extract the meaning of a sentence. It labels words or phrases that specify their semantic role. [Palmer2010]Semantic roles include agent, themes, goal, instrument, source and result, etc. An example of labelling semantic roles is, Raman/AGENT broke the mirror/THEME with a hammer/INSTRUMENT.

**Disambiguation:** Also referred to word sense disambiguation, which is used to determine the sense of a word [Navigli; 2009]. It is a process of discerning the meaning of a word in a particular context. It is utilized in speech recognition and unstructured data analysis, etc.

**Markov model:** The Markov Model models dynamic data such as temporal and sequential data. It models the reliance of current information on the previous information [Petrushin; 2000]. Markov property states that the future state is dependent only on the current state, but not on past events. It plays an important role in predictive modelling. In Markov models, the state will be clearly noticeable to viewer, whereas in hidden markov model, state will not be clearly noticeable but the state-dependent output will be noticeable. It is applicable in speech recognition, part-of-speech tagging, handwritten recognition, and gesture recognition, etc.

**Statistical models:** It is a technique to summarize the collected data on the basis of the closest parameters. It uses mathematical equation to resemble reality and make predictions on the generalized data.

**Recurrent Neural Network:** Recurrent neural network [CC Aggarwal 2018] is a subset of ANN, where nodes are connected in the form of a directed graph and depicts temporal property. Unlike neural networks, RNN has inputs and outputs dependent on each other. The prediction of the next word is dependent on the preceding words. It has memory capacity, which remembers the calculated information. It has its application in pattern and image recognition, machine translation, etc.

**Convolution Neural Network:** Convolution Neural Networks are the networks used for analyzing artificial imagery, with the main building blocks being the convolution layer (comprising of independent filter layers) [S Skansi 2018]. Like other neural networks, they are made up of neurons along with their weights, wherein each neuron receives various inputs and takes the weighted sum of the inputs. These are passed onto the activation function and then respond with an output.

**Feature engineering:** It is an art to recognize features using domain knowledge, which accelerates machine learning algorithms [A Zheng 2018]. The process involves various steps such as deliberating features, planning features to be created, fabricating features, scrutinizing the working of features with the model, enhancing features, and lastly investigating more features through brainstorming if needed.

**Sentiment analysis:** Sentiment analysis, also known as opinion mining, is a subset of text mining, which intends to extract intuitive information [CC Aggarwal 2018]. It aims to understand the opinions, emotions, feelings expressed in a text by classifying them as positive, negative or neutral. It is applied in recommender system to predict the future growth of a particular item, social networking or e-commerce sites to extract users' reviews for a particular item.

**Paraphrasing:** Paraphrasing is restating text using different words, without altering its meaning [I Androutsopoulos 2010]. It is used to understand the meaning of a text in different context and is performed by using appropriate synonyms, changing the order of words, and altering the grammar, etc.

**Semantic relations and its types:** Semantic relations [G.A. Miller 1995] denote how the terms are semantically related to another term. There are various types of word relationships: synonyms, antonyms, hypernyms, hyponyms, homonyms, meronyms, polysemy, etc.

- **Synonyms** denote the association between words having similar meanings. Examples of such words are happy/cheerful, love/affection.

- **Antonyms** describe a link between words having different meanings. They are contradictory in nature. For example, boy/girl, old/young, happy/unhappy, on/off.

- **Hyponyms and hypernyms** are the relations connected in a hierarchical relationship in which the hypernym is generic and superordinate to the hyponym. For example, lion, tiger, dog belongs to animals, thus, animal is the upper term called hypernym, while lion, tiger, dog are the lower terms called hyponyms.

- **Homonyms** relate words with distinct senses. They are of two types: **homophones** connect words with a similar pronunciation, but different spelling and meaning, e.g., sole/soul, son/sun, steal/steel, stair/stare, while **homographs** connect words having the same spelling, but different meanings and pronunciation, e.g., *bass,* which means type of fish, or a low voice; *bow,* which means type of knot or to incline.

- **Meronymy** illustrates part-whole relationships between two words, which means if A has B then B is a part of A. Examples are: collar is a meronym of shirt, cover and page are meronyms of book.

- **Polysemy** consists of poly (many) and semia (having meaning). It is an expression having multiple meanings, but they are conceptually related to each other. For example, the verb *get* means understand, become, acquire.

**Statistical translation:** Statistical machine translation translates text written in one language to another language [P Koehn 2009]. The probability of translation is denoted as p (e|f) where 'e' is a translation of 'f' where string 'e' is written in the target language, while 'f' is a string written in the source language; this helps to determine the translational-probability of a document

**Excitation:** This semantic property is used in causality that categorizes templates into excitatory, inhibitory and neutral. Excitatory templates enhance the effect, role, and functionality of the referent, e.g. enable Y, produce Y, increase Y; inhibitory templates suppress the effect, role, functionality of the referent, e.g. disable Z, decrease Z, prevent Y; neutral templates neither enhance nor suppress e.g. consider Y, evaluate Y, related to Y etc. [Hashimoto et. al. 2012]

**Support Vector Machine:** A support vector machine [S Suthaharan 2016] is a supervised machine learning algorithm that classifies a set of data objects. It outputs a decision hyperplane and divides newly input data objects into two classes lying on either side of the plane by identifying their features.

**Naive Bayes:** Naive Bayes algorithm [KM Leung 2007] is one of the probabilistic classifier in machine learning. It uses Bayes Theorem with naive conditional independence between the attributes.

**Prediction:** Prediction is a data analysis technique that builds a model for continuous ordered values and predicts future trends, while classification predicts discrete labels.

**Regression:** Regression analysis is a statistical method that predicts numerical data and models [DC Montgomery 2012]. It helps in the understanding of the relation between mean values of a variable and values of other related variables.

## 4. General Architecture of Question Answering System

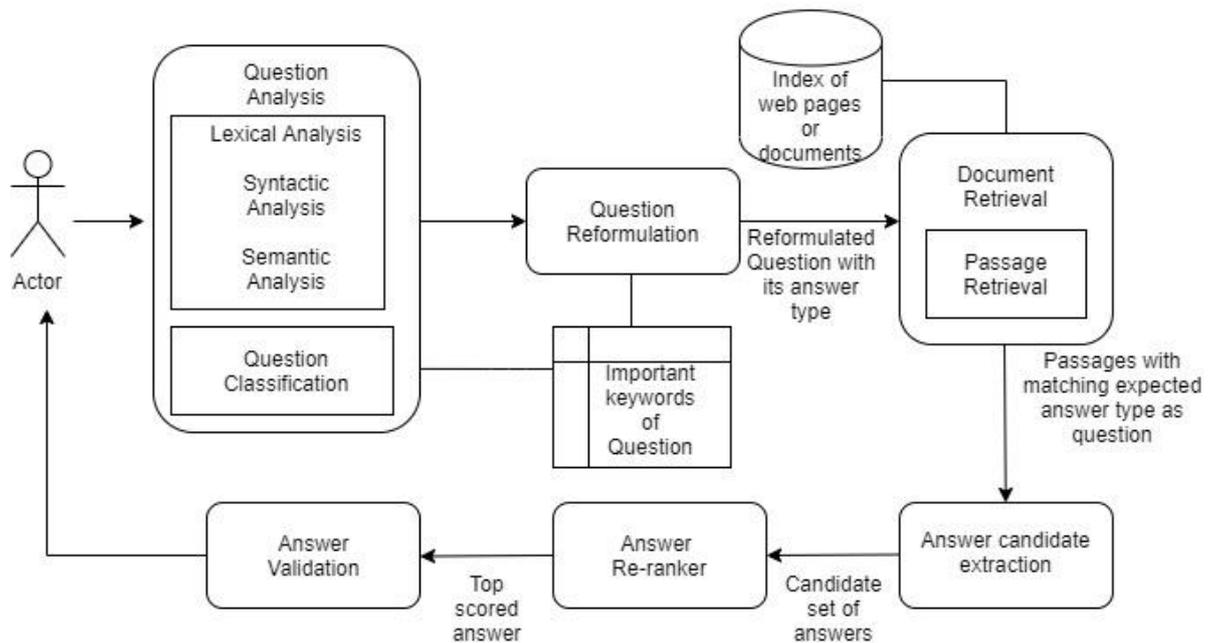

*Figure 1: Architecture of Question answering System*

Question Answering System comprises of five main modules independent of the type of questions being asked by user, which are Question Analysis and Processing, Document and Sentence Retrieval, Answer Extraction, Answer Re-ranker and Answer Validation. Each module plays an important role in improving the performance of QAS.

The question analysis module processes the input question to look for its focus, type and its expected answer type. It performs lexical, syntactic and semantic analysis of questions and categorizes it, the accuracy of which aids the retrieval of its correct answer. Lexical analysis involves techniques such as tokenizing, POS tagging, keywords extraction, stop-word removal and Named entity tagging. The aim is to break the questions into smallest meaningful terms called tokens and recognize the role of each term. Syntactic analysis aims to determine the syntactic form or order of arrangement of terms in a question. The most common form of this phase is dependency analysis which finds the relationship between two lexical terms contained in a question according to the rules described in dependency grammar. (used to build the structure of phrase and sentence by representing dependency relation between the words used in phrase and sentence). The process is automated using dependency parser which generates parse tree of a question according to the hand-written rules. Now, after counting the lexical terms and looking for their arrangement in a question, semantic analysis phase tries to determine the meaning of words in a question. It uses semantic role labelling that labels words or phrases in a question and assigns them semantic roles like agent, goal or result. It utilizes knowledge base developed by three resources: FrameNet which assigns semantic roles to predicates, PropBank is a corpus comprising of annotated verbal propositions, and VerbNet provides classes to verb types defined in PropBank [Giuglea et. al.; 2006]. Thus, it tries to draw the inferences and needs of user so that the system can extract an appropriate answer to a question.

Further question analysis phase classifies the questions on the basis of syntactic and semantic information and determines its expected answer type. It has been stated that [Moldovan et. al.; 2000] the results of question classification affects the performance of QAS. According to the authors in [Mishra; et. al; 2016], QAS are classified on the basis of various factors like (1) Application domain: General and Restricted domain (2) Types of questions: Factoid, List, Hypothetical, Causal, and Confirmation questions (3) Types of analysis done on Questions : morphological, syntactic, semantic, pragmatic and discourse (4) Types of data sources: structured, semi-structured and unstructured (5) Types of matching functions used in retrieval models: Set theoretic models, Algebraic models, Probability models, Feature based models, and Conceptual graph based models (6) Characteristic of data sources: source size, language, heterogeneity, genre, and media (7) Techniques used in QAS: Data Mining, Information Retrieval, Natural language understanding, Knowledge retrieval & discovery and (8) Forms of answer: Extracted and Generated answer.

After the question classification module, question analysis module aims to reformulate the question with techniques like relevance feedback and query refinement to showcase the user's appropriate need. The reformulated question is input to the search engine, which returns a list of documents. These documents are ranked according to their relevance to a query which is a combination of various metrics like count of keywords, hit ratio, user history logs etc. Passage retrieval module divides the top ranked documents into smaller passages like paragraphs and sentences using various paragraph segmentation algorithms. These passages are retrieved on the basis of various features like number of named entities, keywords, order of keywords in a question as well as the passage and answer type determined from the passage.

From the retrieved passages, the answer extraction module draws out relevant answer candidates that match the answer types returned by the question analysis module. Some features described above in passage retrieval also affects the answer candidate retrieval like (1) containing phrases matching with the correct answer type, (2) number of keywords match with question and answer candidate, (3) novelty factor in candidate answer, (4) sequence and arrangement of words in question and candidate answer,(5) location of punctuation in answer candidate.

The answer re-ranker module re-ranks the extracted answer candidates. The most appropriate answer with the highest score is delivered as an appropriate answer output to a user.

5. **Techniques involved in modules of Why-type QAS**

This section discusses the existing as well as proposed techniques which are used in the development of various modules of QAS that can answer Why-type questions.

5.1. Dataset Preparation

The development of Why-type QAS requires a dataset of Why-type questions. Questions are collected from resources depending on the requirement of QAS, whether it is open-domain or restricted-domain. There are various resources available online such as Yahoo! Answers [Yahoo url], Answers.com [url], Quora, WikiAnswers, AskMe etc. from which Why-type questions are extracted i.e. questions starting from Why are filtered. Question Answer (QA) may be extracted from the tracks carried out in various conferences/workshops relating to Information Retrieval. One such collection is TREC GOV2, which contains more than 25 million documents, which can be used to extract why-QA [Clarke et. al.; 2004]. Some other sources include but may not be restricted to FAQs, Webclopedia QAS (which extracts answers from Wikipedia text documents) and Microsoft RFP collection of queries posted on the Microsoft Live search engine. In our research [Manvi 2018], we have prepared a dataset of 1000 open domain Why-type questions from Yahoo! Answers, Answers.com, WikiAnswers, Twitter, Quora and Suzan Verberne [Verberne url] website by filtering questions beginning with Why.

Whenever text documents as knowledge source are provided, from which questions and answer passages are to be formulated, researchers take the help of human annotators to prepare a question-answer dataset from the appropriate text passages. Annotators formulate the questions and their answers from the given document by utilizing his/her knowledge and experience. Dataset is also enlarged by paraphrasing the questions formulated, where the questions are restated using different words without altering their meaning. This method yields the better results if the annotators involved have better domain knowledge and similar experiences.

Particularly for Why-QAS, causal relations play an important role in developing QAS. Questions and answers are represented by cue phrases which are used to connect two sentences, serving as a cause and its effect. Jong-Hoon Oh et.al. in 2013 [Oh 2013] used causal phrases like 'because', 'this causes', 'caused by' and employed the causality recognition method to extract causal relations from web texts, which further aid the generation of Why-QA pairs. He asserted that the effect component in causal relations is the foundation of generating questions, whereas the cause component seems to be an answer to that question. For example, [Tina is absent]$_{effect}$ because [she is sick]$_{cause}$. Here, 'she is sick' formulates an answer to the question 'Why is Tina absent?'. This method shows positive results in case of explicit cue phrases involved in Why Questions but it doesn't perform well in cases of implicit cue phrases where semantic knowledge is required to extract QA pairs.

| Challenges | Solutions |
| --- | --- |
| 1. Questions posed on social QA websites are somewhat grammatically incorrect/ no valid meaning | Automatic pre-processing of questions needed which scans them for grammatical errors/spelling/order of terms |
| 2. Collecting/Preparing dataset for Why-type questions is cumbersome | Need of crawler that crawls websites and collects Why-type questions |
| 3. Why questions are not limited to those starting with Why | Need of a tool that identifies questions involving causal component, not only questions that start with Why |

*Table 2: Challenges in Dataset Preparation and Future Directions*

## 5.2. Question Analysis and Processing Module:

Question Analysis module aims to extract important keywords to understand the need of user which can help determining the approach of answering it. This process comprises analysing question lexically, syntactically, and semantically followed by question classification. Different techniques exist for processes carried out in the Question Analysis stage such as tokenization, word disambiguation, POS tagging, entity annotation, logical forms, dependency parsing, semantic role labelling, and co-reference resolution etc. In this phase, questions are parsed through syntactic dependency parser to deduce the question structure and formulate the semantic patterns by utilizing semantic information contained in it. After understanding the questions, it is input to the question classification module which classifies the question according to the keywords and determines its expected answer type.

Suzan Verberne in 2006 [Verberne 2006] has implemented syntactic analysis for keyword extraction. She has recognized named entities by applying named entity recognition and extracted noun phrases by shallow parsing. For example, the question 'Who is the Prime Minister of India' translates to query 'Prime Minister India' after removing stopwords like Who, is, the, of. But this process had certain shortcomings as the query formed by the keywords is not the most appropriate in retrieving documents. In such cases, the query needs to be reformulated by appending relevant terms according to the user's need. This plays a major impact in answering Why-type questions as the documents which are extracted by matching keywords may not always fulfil the user's need, because answers to Why-questions require reasoning behind the occurrence of the fact rather than merely retrieving information of the fact asked in a question. For example, a why-question "Why do people cry?", the documents retrieved may contain the information relating to the description of the phenomenon "crying", and its process. The specific need of the user to find an answer explaining the reason behind the crying act can be found by scrutinizing the documents containing this information.

### 5.2.1. Question Classification:

'Question Classification' is the second phase of Question Analysis module which categorizes the question syntactically or semantically by utilizing important keywords extracted in order to probe its expected answer type. Many researches have proposed a lot of taxonomies based on different factors for factoid type QA while limited work has been done in non-factoid QA. Some of the researches carried out in proposing the taxonomy for Why-type questions are 'Per Holth' in 2013 [Holth 2013] who did behavioral analysis of Why-question. The author proposed different categories such as *Immediate Antecedent*, which aims at knowing the immediate predecessor of the event; for example, 'Why did the window pane break?', *Disposition or Summary Label* asking reasons under circumstances in which the event occured; for example, 'Why did the window pane break when hit by ball?', *Internal Mediating Mechanism*, which covers Why questions asked in neurology that queries the reasoning behind the inner workings of the event; and *External Historical Variables* which questions the behavior of the subject under the influence of external variables. Ferret et. al.

[Ferret 2002] had parsed a question with the use of a shallow parser and applied handwritten rules to the resultant parse tree to find out the syntactic category of the question. Suzan Verberne; 2006 [Verberne 2006] has utilized Ferret's approach to syntactically categorize why-questions as *action questions*, *process questions*, *intensive complementation*, *monotransitive have questions*, *existential there*, and *declarative layer* questions. According to Moldovan et. al., all why-questions possess *reason* as its answer type. However, Suzan Verberne surmises that there is a need to split the answer type (reason) into different subtypes to select the appropriate sentences. Thus, she proposed subtypes of reason on the basis of adverbial clauses given by Quirk [2010] as *cause, motivation, circumstance, and purpose*.

Motivated by the importance of question classification for improving the performance of question answering system, we have tried to extend the research on proposing a taxonomy of only Why-type questions. A classification was proposed after analysing a dataset of around 2000 Why-type questions. It was proposed by analysing POS tags, conjunctions and other set of key terms used in the question. Following four categories were proposed namely We have extended the research on classifying Why-questions [Manvi; 2017, 2018] and proposed four categories *Informational Why-questions,* which require an explanation of the facts in their answers; *Historical Why-questions,* which provide the justification for the events that have happened in the past; *Contextual/Situational Why-questions,* which give the reasoning behind the events that have occurred at a particular time; and *Opinionated Why-questions* asking interpretations for the person or some entity, depending on the individual's knowledge and experience. Further, we have determined expected answer types corresponding to a given question on the basis of lexical words contained in a question as discussed in section 5.4.1. We also plan to explore patterns contained in different answer types that help to find a relevant and highly ranked answer to Why- question.

Besides the syntactic approach for why-question analysis, Karyawati in 2015 [Karyawati; 2015] adapts the Bag-of-word model with semantic entities to represent a query which captures user needs. The researchers used a methodology combining POS tagging with typed dependency parsing to construct the patterns of why-questions which depict the relations between the terms. Verb classification and domain ontology was employed to determine the expected answer types. The method showed good performance measures but drawback being excessive time consumption for manual construction of lexico-syntactic patterns and resultant generation of limited number of patterns which could not address all real question patterns. Further, the research was implemented with the assumption that why-questions must be in correct English grammar and address specific domain i.e. text-retrieval domain. Since the above method showed significant improvement which can further could improve other baseline methods (explain), it can be improved by automatically generating lexico-syntactic patterns using machine-learning techniques and expanding the domain ontology by constructing the semantic index.

| Challenges | Solutions |
|---|---|
| If lexical patterns are handcrafted, it cannot cover a range of questions, i.e. limited patterns are possible | Automatic generation of patterns which are formulated with different features using machine learning techniques |
| Need to determine user focus and expected answer type | To improve performance, semantic features need to be employed with domain knowledge |
| Ambiguous category assigned to some questions | Appropriate weighting of features required that can resolve ambiguity and assign a category to given Why-question |

*Table 3: Challenges in Question Classification and Solutions*

5.2.2. Question Reformulation:

After Question analysis, there is a need to reformulate the questions according to its class defined by question classification. This is required to depict an appropriate user need from the query that helps to extract an appropriate document to a query. Question Reformulation forms the bedrock of a question processing module, which reformulates the question by incorporating important terms to enhance the understanding of the user requirement. According to Carpineto and Romano in 2012 [Corpineto; 2012], various query reformulation techniques comprising syntax relations, semantic relations, and usage knowledge (explain) are utilized to choose the important terms. The original question (Q) is expanded with the terms contained in answer documents to avoid large vocabulary mismatches between question and answer candidates. Various approaches have been utilized like (1) expanding query with semantic relations found in WordNet (Miller; 1995). Terms are figured out from WordNet that depict lexical relations between question phrases and their answer documents. (2) Query is refined by interacting with user in which opportunity is given to them to pick appropriate terms to suit his/her requirements (M. Harvey; 2015). (3) The system suggests relevant queries to user that he can choose as expanded terms. This is commonly viewed in Google Assistant where Google search engine assists the user with a list of relevant queries (Harvey; 2015). (4) Terms to be appended to query are chosen from query expansion term space (QETS) which is selected by finding the proximity between their outlink pages. Ganesh et. al. in 2009 [Ganesh 2009] explored the content and structure of Wikipedia pages for query expansion. Semantic relatedness of terms are calculated by the summation of the proximity and outlink score, where the proximity score of a term is weighted by its frequency and the minimum distance to a keyword in the question over all relevant sentences (S) in Wikipedia and outlink score exploits the link structure and category information of Wikipedia.

Thus syntactic as well as semantic relation between words plays an importance for question reformulation. The accuracy of this module can be enhanced by referring the user's previous question logs and expanding query with terms used in previous question. Also the results of this module are affected from the domain of QA. Terms to be appended depend on the available and accurate domain knowledge. Sometimes expert knowledge can be used to reformulate the query in order to target question to an accurate answer.

| Challenges | Solutions |
|---|---|
| Understanding the accurate user need from question | Question to be paraphrased or reformulated with lexical terms corresponding to user need which are identified from semantic relations and user logs |
| Question cannot address or identify the domain | Domain and expert knowledge required to reformulate the question that help direct question for appropriate passage retrieval |

*Table 4: Challenges in Question Reformulation and Solutions*

5.3. Document Retrieval

'Document Retrieval' identifies documents that are likely to contain an answer to a question. This module returns relevant documents which contains content related to a query. Documents are deeply analysed on different parameters like *pattern matching* which matches phrases in document as used in question, *syntactic parsing* which utilizes parse tree or dependency graphs to depict the right syntactic position of phrases, *synonym and semantic parsing* which assigns synonyms and semantic types to phrases using named entity recognizer. It matches documents against user queries and evaluates the matching results by sorting them with their relevance score, calculated using a PageRank algorithm. The retrieval engine either returns an unordered list of relevant documents or generates a ranked list of documents that are scored on the basis of their likelihood of containing an answer (Monz C. in 2003).

Passage Retrieval is a subpart of Document Retrieval which aims to identify location of the relevant paragraphs/passages from the retrieved documents. The whole document is scrutinized for a specific paragraph which has a maximum probability of containing an answer. These passages are selected by identifying an overlap of words in a query and passage. This technique was used by Suzan Verberne [Verberne; 2006] in her research. She retrieved documents by matching query words with answer candidates using Bag of words model. N-grams from query and the documents are retrieved and are matched. They are weighted using a tf-idf scoring where tf is the number of instances of terms in the query and the relevant passages and idf is the inverse document frequency that weighs each document with the instances of terms in it. The model was applied to Why-QA and faced various shortcomings in cases of short questions containing only one semantically rich content word, the correct answer document may be listed much lower in the retrieved set of documents because of lack of understanding document context, and multi-word phrases which are considered a single term to retrieve appropriate documents.

This model doesn't give the best results as Why-QA demands the understanding of terms within the question to extract the appropriate document rather than matching terms with the document. Since this model doesn't consider the order of terms, it ignores the context and in turn the meaning of the words in the query and document (semantics), whereas an ideal Why-QA retrieves the correct answer, when what the user desires from the question is properly understood. Thus, to some extent, this model works by retrieving such documents that talk about the terms contained in the question but this doesn't ensure that the appropriate document will be retrieved that actually contains the reasoning behind the question.

Thus, to improve the performance of passage retrieval module, various semantic relations such as "part-of", subset and is-a etc. and inference rules (explain through examples)can be used. Each paragraph is to be weighted on the basis of average scoring from different features and will be given as input to answer candidate retrieval module.

| Challenges | Solutions |
|---|---|
| Different lexical terms used in question and corresponding accurate documents | Need of searching relevant documents with actual question and paraphrased questions |
| Not accurate documents retrieved because correct documents contain same words but not address the need of user | Understand the user needs from the question to search the documents in that direction rather than searching documents from overlapping words |

*Table 5: Challenges of Document Retrieval and their solutions*

| Literature | Methodology and Module | Techniques/Models | Comments |
|---|---|---|---|
| Verberne 2006 | Keyword Extraction (Syntactic Analysis) | Named Entity Recognition | Query formed by keywords may not be appropriate to retrieve documents |
| Per Holth 2013 | Question Classification | Proposed classification based on behavioral analysis of questions | Highlighted confusion in accepting varied explanations for similar kind of why-questions |
| Ferret 2002 | Question Classification | Applied Handwritten rules on Parse Tree | Yielded syntactic category of questions, without describing mapping between syntactic categories and answers |
| Verberne 2006 | Question Classification | Extended Ferret's research to classify questions into action, process, montransitive 'have, intensive complementation, existential 'there' and declarative layer | Based on syntactic categorisation and answer type determination |
| Manvi 2018 2017 | Question Classification | Informational, Historical, Contextual/Situational,Opinionated | Classified on a small dataset. Ongoing refinements |
| Miller 1995 | Question Reformulation | WordNet | Links various parts of speech to sets of synonyms |
| M Harvey 2015 | Question Reformulation | Interactive Query Refinement | Systems showing suggestions to enhance user query, can lead to better user queries |
| Ganesh 2009 | Question Reformulation | Relevance Feedback | Query expansion to rank answer containing passages better |
| Monz C 2003 | Document Retrieval | Stemming, Blind Relevance Feedback, Passage Based Retrieval | Document scoring on chances of containing an answer |
| Verberne 2006 | Document Retrieval | Bag of Word Model | Match query words with answer candidates. Approach ignores context which may not yield best results |

*Table 6: Overview of the existing research on Question Analysis and Document Retrieval modules.*

5.4.  Answer candidate extraction:

This module extracts the relevant set of answer candidates from the documents or passages retrieved in the previous module. For a why-type question, there may be multiple answer candidates. The appropriateness of answer candidates retrieved affects the final accurate answer. Techniques that help the extraction of answer candidates in Why-QA are discussed in brief below:

   5.4.1.  Lexical-Syntactic Analysis:

Passages are analysed lexically and syntactically which helps identify the terms and extract the meaning of text by tagging the tokens with parts of speech such as noun, verb, and adjective. In order to understand the relations between the entities involved in a sentence, Suzan Verberne in 2007 [Verberne; 2007] had utilized discourse structure for answer extraction from retrieved documents. She used Rhetorical Structure Theory (RST) as a model for discourse analysis that finds rhetorical relation between two text spans (https://www.sfu.ca/rst/). Since answers to why-questions provide reasoning to an event asked in the question, various relations pertaining to causal relations such as cause, purpose, motivation and circumstance have been identified in the passages. The sentences pertaining to such relation types as found in questions, are extracted as answer candidates. Manual analysis was done on 336 QA pairs out of which for 195 why-questions (58.0%), the correct reference answer was found. For about 141 questions (42.0%), answers could not be extracted in document collection. Such questions are distinguished as:

(1) Question for which sufficient world knowledge is required to find the correct answer for ex. The text fragment "Cyclosporine can cause renal failure, morbidity, nausea and other problems" can deduce correct answer to the question "Why is cyclosporine dangerous?" if we know sufficient knowledge regarding renal failure, morbidity, nausea and other problems to be dangerous.

(2) For 16.4% of the questions, the relevant answer candidate contains both question topic and answer but no RST relations corresponds between the two spans.

(3) For 5.1% of the questions, the answer's location couldn't be found from the text although question topic is supported.

(4) For 3.6% of the questions, nucleus part of RST relation matches with the question topic but its corresponding satellite part doesn't depict correct answer. There is a need to automate RST annotations but it is less complete and precise than manual annotations. Thus, partial automatic discourse annotations can be employed where it is feasible to provide some information needed for answering why-questions. The method fails in the cases where there are no explicit relations or causal patterns (e.g. tidal waves can be caused by earthquakes, I got late because I was stuck in jam etc.) found in the answer text but implicit causality is involved (e.g. cold tremble, malaria mosquitoes etc.)

In our research, we have also used Rhetorical Structure Theory (RST) relations [Manvi; 2018] as a ground to proffer different classes of answer types expected for Why-Questions as *comparative answer* describing the comparison of facts, *motivated answer* containing motivations behind the actions, *conditional answer* containing reasons under a different time context, *justified answer* providing reasons for the inventions to be proved by theory, *unconditional answer* describing the cause of some event irrespective of any circumstance, and *interpreted answer* containing logical reasoning for the facts related to the domains of logic, maths and statistics. RST provides a set of relation names which provides the relation between two spans of text, referred to as nucleus and satellite. Why-QA focuses on providing the reasoning, thus the nucleus part of the text claims the event whereas satellite part of the text provides the evidence to it. There are two types of relations, one existing between nucleus and satellite and other existing between multiple nucleus. We have tried to map these relation names with the expected answer type, assuming the correct answer candidate to a question contains such relations between the text spans which are discussed in the table below

| Expected Answer Type | RST Relation Names |
|---|---|
| Comparative | Otherwise, Unless, Conjunction, Contrast |
| Motivated | Antithesis, Concession, Enablement, Motivation |
| Conditional | Condition, Circumstance |

| Justified | Justify, Evidence, Means, Solutionhood, Elaboration |
| --- | --- |
| Unconditional | Non-volitional cause, Non-volitional result, purpose, unconditional, volitional cause, volitional result |
| Interpreted | Interpretation, Evaluation |

*Table 7: Relation of RST relations with expected answer type*

The answer types are identified by finding the lexical features in a question. Thus, the sentences which best correspond to the features are better represented as an answer candidate to a question.

5.4.2. Causal Relation

Causal relations are closely associated with Why-QA as it describes the explanation in their answers. Various techniques have been adopted by researchers to identify causal relations between question and passages to extract answer candidates.

The research initiated with finding the hand-written causal patterns in the text, and answer candidates are retrieved which contain those appropriate patterns. The authors who extracted answer candidates by matching different causation patterns using different approaches and different QA pairs are discussed.

Girju in 2002 and 2003 [Roxana Girju; 2002, 2003] presented a learning approach for automatically discovering lexico-syntactic patterns exhibiting causal relations and proposing a taxonomy of causal questions. The authors considered intra-sentential pattern of the form <$NP_1$ verb $NP_2$> which is discovered by choosing a causal semantic relation e.g. CAUSE-TO and choosing noun-phrases that holds the relations. They also elucidated question classes as, (1) explicit causation questions containing exhaustive keywords as, effect, cause, consequence etc., (2) ambiguous or semi-explicit questions containing exhaustive and ambiguous keywords reflecting causation relation which when recapitulated determines its semantic type as create, trigger, produce etc. and (3) Implicit questions don't use explicit keywords but implicate reasoning with deep semantic analysis, common sense and background knowledge. However the system lacked due to (1) ambiguity of causal relations (e.g. trigger, lead to, elicit, originate etc.), (2) limited named entity recognition (e.g. the names of person, places, animals are not exhaustive) in WordNet and (3) research considered only causal patterns encountered within the sentence (e.g. Tsunamis are generated because ocean's water mass is displaced), except the patterns across the sentences (e.g. Earthquake causes seismic waves. This causes tsunami). This ignited the need for automatically detecting causal relations for answering Why-type questions.

Hovy in 2006 [Hovy; 2006] had proposed Basic Element (BE) method that elucidates semantic relation between two elements which is utilized by Fukumoto in 2007 [J Fukumoto; 2007] to provide different answer extraction patterns for non-factoid type questions (why-type, definition-type and how-type). The authors analysed various extraction patterns of the form (Verb + because, Noun + because etc.) and non-extraction patterns of the form (Pronoun + Postposition + because, Verb + because + Postposition etc.). Answer candidates are marked relevant if they match the extraction patterns and those which resemble non-extraction patterns are removed. The method attained 50% accuracy when compared with scoring provided by human. The method gave limited results owing to number of question and answer extraction patterns. BE scoring fails for those answer candidates whose syntactic structures are different but their meanings are same. Thus, the performance can be improved by considering paraphrasing on answer candidates which make use of different words but their meaning remains the same.

Pechsiri and Kawtrakul in 2007 [Pechsiri; 2007] identified causality events and areas of causative and effective units in a document by extracting causality knowledge within numerous EDUs (Elementary Discourse Units). EDUs are the minimal discourse unit which are represented by the leaf of RST tree corresponding to a text. EDUs are the short sentences or clauses which comprise multiple causes and their effects. The method was applied on the texts related to agricultural and health domains from which inter-causal and intra-causal EDUs were exploited by learning verb-pair rules using SVM and Naïve Bayes classifier. The approach performed well with accuracy of 96% in agricultural or health news domain, however it faced limitation in identifying effective boundary between successive EDUs. It can be utilized in other domains by representing causality with different discourse markers and NP pairs faces which gain high precision by extracting from WordNet. The methodology achieved higher accuracy of 96% but disruption between successive EDUs challenge effective boundary identification. Also despite considering verb and NP pairs, other parts of speech can also be utilized.

Mori et. al. in 2008 [T. Mori; 2008] proposed a novel method of answering non-factoid Japanese questions. They formulated Q&A pairs from the collection of questions asked on social Q&A website. The focus of each questions are identified by keeping the set of functional & content words like interrogatives, verbs, adjuncts, reason, method etc. but replacing other words with their parts-of-speech. Appropriate answer candidates are identified by scoring the sentences on two measures where Measure 1 corresponds to finding the content similarity between a question and answer candidates and Measure 2 corresponds to mapping lexico-syntactic patterns and clue expressions from previously extracted answers of some questions having same writing style. The methodology was performed on different Why, how and definition type Japanese questions and clue expressions combined with topical content information improved the accuracy which can further be enhanced by incorporating advanced language model and scoring function.

The above methods focused on identifying causation patterns from the text which helped extract appropriate answer candidates. Further, researches discussed below addressed causality in Why-QA by identifying cause and effect parts in a sentence using different techniques.

Reyes S. and Elizondo in 2012 [S Vazquez-Reyes; 2012] proposed a methodology to answer causal questions with the prime goal of understanding the question. Features consisting of bag-of-words, syntactic and lexical semantic were utilized. Since an answer to why-question entails opinions, interpretations or justifications, causal relations is required to identify opinions and answer causal questions. The answer extraction is based on four measures, which are simple matching (finding total weight by summing the weights assigned to stop words and non-stop words appearing in an answer), longest consecutive subsequence (measures the presence of consecutive words in the question and possible answer), Sorensen's similarity coefficient (calculates the similarity between question and answer text using Sorensen index which is the ratio of twice the number of elements common to both sets to the sum of number of elements in each set), and WordNet-based Lexical Semantic Relatedness (using WordNet as a resource to find similarity between question-word with text-word, synonyms of text-word and synonyms of text-word antonyms). They are weighted accordingly to rank candidate answers.

The research bestowed knowledge effectively to extract features and classifies why-question. The method achieved recall of 36.02% and MRR of 0.219 on texts containing explicit and ambiguous answers. In case of text containing implicit answers, recall was measured as 61.46% of former and MRR of 0.373.

To improve the existing work, the authors claimed to go beyond lexico-syntactic approach and experiment with techniques representing knowledge and reasoning which aims for addressing knowledge, justification, and the rationality of belief behind the answers to why-questions.

Jong-Hoon Oh et. al. in 2012 [Oh; 2012] used both intra and inter-sentential causal relations between terms or clauses to answer why-questions, where the effect part of the sentence depicts the question asked and its cause part provides an answer to a question for ex. The inter-sentential causal relation represented as [Earthquake causes seismic waves which set up the water in motion with a large force]$_{cause}$. This causes [a tsunami.]$_{effect}$ and intra-sentential causal relation represented as [Tsunamis]$_{effect}$ are caused by [the sudden displacement of huge volumes of water.]$_{cause}$. In both sentences, effect part depicts the question "Why are Tsunami generated?" and the cause part forms an answer candidate to a question. The causal and effect part is identified by the causal relation which separates the two parts in a sentence. The methodology was applied on Japanese why-type questions and causal relations are restricted to cue phrases as *"because"*, *"this causes"*, *"are caused by"* and *"as a result"*. Contextual features also play important role to retrieve an appropriate answer candidate for which cue phrases were identified in answer candidates using regular expressions, and for each cue phrase, the authors extracted three sentences, one containing phrase, other two are its preceding and succeeding sentences in answer candidates. The appropriateness of causal relation as an answer was also measured using term matching (where the effect part must contain atleast one matching content words like nouns, verbs, adjectives as in the question), partial tree matching matching (where the effect part must contain atleast one partial tree which covers more than one content words as in the question), and excitation polarity matching (meaning is captured by identifying whether the role of noun/entity in text is activated or suppressed). Thus causal relations and excitation approaches are utilized for finding appropriate answer candidates and thus proved that the system could achieve 83.2% precision for its appropriate answer candidates.

Jong-Hoon Oh in 2016 [Oh; 2016] extended his previous work and retrieved passages consisting of seven sentences having atleast one cue phrase which is used for recognizing causal relations. Thus, from a large collection of 2 billion web texts, the method extracted about 4.2 billion passages. To retrieve appropriate answer candidates, two types of Boolean queries are generated e.g. "$n_1$ AND $n_2$ AND .. $n_j$", and "$n_1$ AND $n_2$ AND .. $n_j$ AND ($va_1$ OR….$va_k$)". Also, it was observed that an accurate answer candidate must contain all the nouns in why-question which helped to retrieve top passages from combined the results of each of the queries. The retrieved candidates are passed to answer re-ranking module.

Jong Hoon Oh et. al. in 2017 [Oh; 2017] extended their work by dealing with implicitly expressed causalities. Since the texts with implicit causality may be expressed in other texts with explicit cues, why-QAS was improved by automatically recognizing explicitly expressed causalities and using them to complement implicitly expressed causalities in answer passage. This was implemented by multi-column convolutional neural networks with causality attention. In this method, archive causality expressions (CEs) (causality expressions automatically extracted from a text archive) were automatically extracted, inner-passage CEs (inner-passage causality expressions) were extracted from a given answer passage and from this, relevant CEs were extracted that are most relevant to both the input question and its answer passage. Causality-attention (CA) words are common words which are extracted from archive CEs, which directly or indirectly associates with the causality between questions and their answers. Thus, different features from questions, answer passages, and causality expressions from answer passages were recognized to associate the words and their contexts in answer passages. The method achieved better performance and effectively improved the quality of the top answers because of the two key parameters viz. causality attention (attention to common words) and relevant CEs (expressions relevant to both question and answer passage).

The authors have tried to improve Why-type QA through a series of their researches. From 2012, 2013, 2016 and then 2017, they have used causality as a ground work for answering Why-type questions. From working on intra and inter-sentential causalities to using it for generating QA pairs, identifying explicit and implicit causalities in QAS, the authors have given new light on improving the performance of Why-QAS.

Karyawati et. al. in 2018 [2018] used semantic similarity measure with selective causality detection for extracting answer candidates. The selective causality detection is applied because only those sentences that contain causality are considered. Sentences are scored using a scoring function which calculates the semantic similarity measure between semantic labelings of question and answer sentences. The sentence is considered relevant if its scoring value is greater than threshold value. Semantic annotations of a question are composed of three sets, which are original semantic annotations (identified from original question), additional semantic annotations (identified from expanded query) and causality annotations (identified from causality expressions in question). Semantic annotations of sentences are applied on the basis of considered text-retrieval domain and ontology schema is used to measure semantic similarity. The method was compared with various baseline approaches and results show that it outperforms in MRR by 16%, P@1 by 15%, P@5 by 14%, Precision by 14% and Recall by 19%. Besides the good performance, it faces performance overhead as it involves semantic similarity which is estimated by using time consuming repeated SPARQL query processing. To reduce this overhead, semantic similarity measure can be estimated using other shortest path algorithms using Dijkshtra rather than SPARQL query processing.

5.4.3. Semantic and contextual analysis

Yang et. al. in 2016 [Liu Yang; 2016] employed semantic and contextual features for retrieving answer candidates for non-factoid Web questions to resolve lexical chasm problem between question and answer terms. Three semantic features were observed (1) Explicit Semantic Analysis [T Gottron; 2011] which employs Wikipedia to represent text using ESA representations where semantic relatedness is calculated by finding cosine similarity between ESA vectors of question and answer sentence, (2) Word Embeddings which represent question and answer words by vector using bag-of-words and skip-gram model where similarity between two vectors is measured by average pairwise cosine similarity between two vectors, and (3) Entity Linking feature which represents queries and answer sentences semantically using entity linking system (Tagme) that obtains related concepts by linking texts to an appropriate knowledge base where Jaccard similarity is calculated between Wikipedia pages linked to query q and sentence s as:

TagmeOverlap(q,s)= (Tagme(q) ∩ Tagme(s)) / (Tagme(q) U Tagme(s)).

In addition to three semantic features, context features are also employed to capture the context of an answer candidate where context is defined by the preceding and succeeding sentences with respect to a given sentence.

The authors experimented and evaluated the effect of semantic and context features, and found that both play an important role and thus enhance the performance of the system.

The performance of retrieving non-factoid answer candidates can be extended by more enhanced features such as syntactic and readability. To retrieve correct answers to Why-type questions, there is a need to determine the accurate meaning for both question and candidate set of answers. Also, some questions need answer related to the context at which it has been asked. Thus, both semantics and contextual requirements are necessary for finding correct answer out of candidate answers.

| Literature | Methodology and Module | Techniques/Models | Comments |
|---|---|---|---|
| Verberne 2007 | Answer Candidate Extraction (Linguistic) | Rhetorical Structure Theory | Used syntactic information to extract answer candidates |
| Karyawati 2015 | Answer Candidate Extraction (Linguistic) | NLP based text mining | Utilise domain ontology and expected answer types |
| Girju 2003 | Answer Candidate | Causal Relations | Automatic detection of causal relations and taxonomy of why-questions |
| | Extraction (Causal Relation) | | |
| Shima and Mitamura 2007 | Answer Candidate Extraction (Causal Relation) | Answering Non-Factoid Question in Japanese using JAVELIN | One Sentence Assumption: Answer within one sentence |
| Fukumoto 2007 | Answer Candidate Extraction (Causal Relation) | BE based evaluation of answers to non-factoid questions | Used handcrafted patterns to match answer candidates. Approach suffered from limited question and answer patterns |
| T Mori 2008 | Answer Candidate Extraction (Causal Relation) | Answering Non-Factoid Question in Japanese | Used Q&A Pairs from social websites |
| Pechshiri and Kawtrakul 2007 | Answer Candidate Extraction (Causal Relation) | Extracting causality knowledge within various EDU's (Elementary Discourse Units) | Performed well in agricultural and health domain which contains verb-pair rules |
| Higashinaka and Isozaki 2008 | Answer Candidate Extraction (Causal Relation) | Machine Ranking algorithms such as Rank Boost and Rankings SVM | Overcome low coverage by earlier handcrafted patterns approach for Answer extraction |
| Reyes S. and Elizondo 2012 | Answer Candidate Extraction (Causal Relation) | Causal Relations using BOW, Syntactic and Lexical Semantic approaches | Aimed to develop a system that identifies and organizes opinions in a question |
| Karyawati 2018 | Answer Candidate Extraction (Causal Relation) | Selective Causality Detection and Semantic Similarity Measure | Application in Text Retrieval Domain. The method is time consuming as it uses repetitive SPARQL query |
| Jong-Hoon Oh 2012 | Answer Candidate Extraction (Causal Relation) | Answering Why-Type Japanese questions | Use of intra and inter-sentential causal relations between terms or clauses to answer why-questions |
| Jong-Hoon Oh 2016 | Answer Candidate Extraction (Causal Relation) | Semi-Supervised Learning | Improved results over previous methods proposed by these authors |
| Jong-Hoon Oh 2017 | Answer Candidate Extraction (Causal Relation) | Multi-column neural networks with causality attention | Automatically recognise explicitly expressed causalities and using them to complement implicit causalities in answer passage. Improvement over earlier approaches because of causality attention |

| | | | and relevant causality expressions |
|---|---|---|---|
| Soricut and Brill 2006 | Answer Candidate Extraction (Statistical Translation) | Unsupervised approach to prepare 1mn QA pairs from FAQ's | First approach towards answering FAQ-type questions |
| Berger 2000 | Answer Candidate Extraction (Statistical Translation) | Analyses the role of statistics to find answers to a question | Propose machine learning approaches using Usenet FAQ and Call-Center dialogues |
| Liu Yang 2016 | Answer Candidate Extraction (Semantic and Contextual Analysis) | Explicit Semantic Analysis, Word Embeddings and Entity Linking Feature | Resolved lexical chasm problem between question and answer terms |

*Table 8: Overview of the existing research on Answer Candidate Extraction module.*

| Challenges | Solutions |
|---|---|
| Different lexical terms used in question and corresponding accurate documents | Need of searching relevant documents with actual question and paraphrased questions |
| Not accurate documents retrieved because correct documents contain same words but not address the need of user | Understand the user needs from the question to search the documents in that direction rather than searching documents from overlapping words |
| Accurate one answer is difficult to retrieve because multiple answers possible for why-type question | Need one summarized answer that must address all the reasons contained in multiple retrieved answer candidates. Novel integrator and summarize answer tool is to be implemented |
| Different answer may be expected from different user | User log needs to be addressed to consider the different interests of the user |

*Table 9: Challenges of Answer Candidate Extraction and their solutions*

5.5.  Answer Re-ranking

Answer re-ranker module takes the collection of answer candidates obtained from answer extraction module and re-ranks them to return one accurate highest ranked answer to a user. Answers are ranked using classifier which is trained on assigning a score to each answer candidate based on a defined set of features. Final score is calculated by summing the scores given to each answer candidate based on the features found in them and thus the performance of the module depends on the features chosen for scoring.

5.5.1. Features related to Bag of Word Model:

Suzan Verberne in 2006 [Verberne; 2006] employed Bag of word model as a technique to re-rank answer candidates in which calculated the overlapping between bag of question items and bag of answer items. Bag of question items contains terms corresponding to noun phrases, main verb, and object contained in the question whereas bag of answer items contains words, verbs in the answer. The overlap function is given as

$$S(Q,A) = (QA + AQ) / (Q+A)$$

where,

QA denotes the count of question terms having frequency more than one in the bag of answer items.

AQ denotes the count of answer terms having frequency more than one in the bag of question items.

Q+A is the total number of terms contained in the bag of question and answer items.

Features are:

i. Term frequency –inverse document frequency scoring:

Tf-idf evaluates the significance of a word with respect to a document collection. Term frequency denotes the cardinality of a term in a document which is calculated as the ratio of number of times a term appears in a document to the total count of terms included in a document, while inverse document frequency denotes the count of documents containing a particular term. This scoring function is used to determine the relevancy of document with respect to the user's query. Term frequency assigns equal importance to each term for example the term 'the' is more frequent than more meaningful terms, which leads to exigency of weighing down the score of documents. The importance of inverse document frequency is to lessen the weight of terms occurring frequently while strengthen the weight of terms occurring scarcely and is calculated as the natural logarithm of the total count of documents to the count of documents containing term. Murata et. al. in 2007 [Masaki; 2007] has used this scoring function to rank answer candidates with respect to the question.

ii. Syntactic structure of the question:

Considering the questions syntactically, there are certain parts of the question which carry a lot of significance to rank answer candidates such as phrase heads, phrase modifiers, subject, main verb, nominal predicate, direct object of the main clause, and all noun phrases. An overlap for those parts of question is measured with the bag of answer terms. Candidates having maximum overlapping parts are ranked higher.

iii. Semantic Structure of the question:

To attain accuracy in answer candidate, semantic features are used to determine the focus and need of question. For most questions, syntactic subject is the main focus but where there is semantically poor subject, verbal predicate is the focus and in case of etymology questions, subject complement of passive sentence is the main focus. Thus, the semantic features such as (1) matching words between question focus and document title, (2) relation between question focus words and all answer words and (3) relation between question focus words and all non-focus words are used for ranking answer candidates.

    iv. Document context of the answer:

Other than the title of document, various other aspects of answer context are also important for answer ranking. For example, the overlap between question terms and Wikipedia document's title, overlap between question terms and caption of answer document, relative position of answer passage in the document, overlap between set of cue words included in section heading and set of words included in section heading of the answer passage.

    v. Synonyms:

There may be lexical mismatch between question and answer terms which are resolved by incorporating the plunk of WordNet synonyms. The features are devised by considering the synonym of the terms contained in the question which comprises the count of question terms having atleast one synonym in the bag of answer terms, and the count of answer terms having atleast one synonym in the bag of question terms.

    vi. WordNet Relatedness:

WordNet is a lexical database that groups nouns, verbs, adjectives, adverbs into synsets that exhibits different concept [Miller; 1995]. These set of synonyms (synsets) are interconnected by various lexical and semantic relations. WordNet relatedness tool is used to calculate the relatedness measures between question and answer terms. The semantic relatedness is calculated by finding percentage of the question terms and their synonyms in WordNet synsets, found in the answer candidate. Other features of WordNet addressing semantic and contextual features also play a great impact for Why-QAS.

    vii. Cue phrases:

The last set of features which have its importance in answering Why-questions is the use of cue phrases. They are the connectives used to depict the relation between two text fragments. Cue phrases such as 'because', 'since', 'as a result' which link causes and their effects help to answer Why-questions.

Although Bag-of-words model forms the basic requirement for extracting best accurate answer by ranking all answer candidates, it has its limitations like it disregards grammatical structure, doesn't consider ordering of words. The authors have proposed features of BOW model for re-ranking answer candidates with the assumption that some parts of question and answer passage affect more for ranking rather than other parts. But the quantitative value of how much each feature effects differ from each other is the main issue, which needs to be resolved for best re-ranking answer candidates.

5.5.2.  Morphosyntactic analysis (MSA):

Another method used to re-rank answer candidates focuses on analyzing them morphologically and syntactically. This approach recognizes n-grams of morphemes, word phrases, and syntactic dependencies from the answer sentence [U Mosel; 2012]. *Morpheme* is the smallest grammatical unit of the word. It can be either a root word having individual meaning or suffix that is appended to other morpheme. For example, a word 'cats' breaks into two morphemes, viz. cat (root word) and s (suffix). *Word phrases* consist of one or more than one words, for example, noun phrase like 'barking dog', verb phrase like 'walking to the door' etc.

*Syntactic dependency* is a binary operation that links denotations of two related words by interpreting the sentence semantically. For example, the sentence 'Mary ran' is connected by 'subj' as syntactic dependency where 'ran' is the head word and 'Mary' is the dependent. More possible syntactic dependencies are "subj" (subject), "dobj" (direct object), "iobj_prep-name" (prepositional relation between a verb and a noun), "prep-name" (prepositional relation between two nouns), "attr" (attribute relation between a noun and an adjective).

This approach is utilized by researches as discussed below:

Jong-Hoon Oh. et. al. in 2013 [Oh; 2013] has utilized morphosyntactic analysis to re-rank answer candidates. Morphological analyzer and syntactic dependency parser is applied on question and answer candidates to extract n-grams of morphemes, word phrases and syntactic dependencies. This helps to formulate four features as (1) count of n-grams of all sentences contained in question and answer candidates, (2) n-grams of answer candidates are found that contains a term of a question, (3) n-grams accomodating a clue term are extracted from respective question and answer candidates, (4) percentage of question terms existed in answer candidate which is calculated as the ratio of total count of question terms in an answer candidate to the total number of question terms.

Unlike BOW model, morphosyntactic analysis considers the grammatical structure and the ordering of words but it couldn't give high performance as semantic features are not included with it. In Why-QA, correct answers could be retrieved when the meaning of question is very well understood and further helps to retrieve the meaning of accurate answer passages which is incorporated by the same authors in their further research.

5.5.3.  Semantic Word Class:

In addition to morphological and syntactic analysis features, Jong-Hoon Oh in 2013 [Oh; 2013] employed semantic word classes for answer re-ranker. Semantic word classes include the collection of words to be used in semantically similar context which are constructed by Noun clustering algorithm proposed by Kazama and Torisawa in 2008 [J Kazama; 2008]. A classifier is trained on a feature representing the associations between semantic word classes present in question and answer candidates. A positive training sample is provided to the classifier which helps it to learn and recognize the correct answer to a question. N-grams of question and answer candidates are converted into their respective word classes and correspondence is found between n-grams used in answer candidate with its semantic word class used in question to find an appropriate relation between them which helps in extracting an appropriate answer to a question. In addition to noun phrases, various combinations of verb, adverb, pronoun, and adjective can be formulated into semantic word classes which will further help to improve the feasibility and effectiveness of the system.

5.5.4.  Sentiment Analysis:

Jong-Hoon Oh et. al. in 2012 [Oh; 2012] introduced sentiment analysis for answer re-ranking in why-QAS in which sentiment polarities of words and phrases included in different answer candidates are recognized. Answer Re-ranker is trained on the features representing sentiment words with distinct polarity. Opinion extraction tool is used to recognize polarity of the sentiment phrases using word polarity dictionary [https://www.kaggle.com/milind81/dictionary-for-sentiment-analysis]. There are two classes of sentiment analysis features, namely word-polarity and phrase-polarity. Word polarity is identified from word polarity dictionary and these features in question answering, aimed at discovering correlations between polarity of words encountered in question and answer candidates. Whereas phrase polarity captures polarities of sentiment phrases included in question and answer candidates. The methodology shows significant improvement in Why-QAS and mainly in opinionated QA where extracted answers reflect the opinion of the words involved in question. Also, for further improvement of QAS, the features can be expanded using semantic orientation, excitatory or inhibitory features.

5.5.5.  Content Similarity features

Higashinaka and Isozaki in 2008 [Higashinaka; 2008] explored various features to train answer re-ranker of why-type QAS based on the likelihood of their contents to be similar. There are three cases, (1) question and candidate answer have identical words, (2) question and answer candidates don't have identical words but still the content is focused and (3) semantically similar content with different sentences. If the question and answer candidates consist of matching terms, they are probable enough to share the same content which is calculated using cosine similarity or n-gram overlap. However, if the question and answer candidates don't contain matching terms but they share similar content found by measuring similarity between question and document containing answer candidates. However if the question and candidate answer are semantically similar but containing different phrases, similarity is measured by utilizing synonyms of words and semantic relatedness features [MAH Taieb; 2013] mentioned in thesaurus of WordNet. Thus, the above approach addresses similarity by comparing the content and focus of question and their answer passages.

5.5.6. Causal Relation Features:

Higashinaka and Isozaki in 2008 [Higashinaka; 2008] presented causal relation features in addition to causal expression and content similarity features to re-rank answer candidates. To overcome the low coverage of handcrafted causal patterns (Fukumoto et. al.; 2007), causal expressions are collected from corpora tagged with semantic relations. Causal relations are extracted from EDR Dictionary [http://www.wtec.org/loyola/kb/c5_s2.htm] and links causal expression found in answer candidate and its subsequent effect in question. Content similarity is mapped using lexical similarity metrics. Authors trained answer ranker module of Why-QAS using machine learning algorithms such as RankBoost and RankingSVM based on the features combining causal expression, content similarity, and causal relation which shows the outperformance of system with 0.305 MRR and high coverage.

The authors have only utilized causal relation as semantic relation, although other relations such as 'purpose' related to causality can also be exploited. Also, the system's performance can be improved by incorporating syntactic and semantic features, and changing their weights while training an answer candidate ranker.

Jong-Hoon Oh et. al. in 2013 [Oh; 2013] utilized intra- and inter-sentential causal relations to answer why-questions. Cue phrases such as 'because', 'since', 'are caused by' between terms in a sentence and 'this causes' between two sentences are utilized to connect cause and effect parts of causal relation. The research by Higashinaka and Isozaki [Higashinaka; 2008] seems improbable due to two challenges, which are identifying accurate causal relations in answer candidates and estimating its aptness.

The authors gauged the aptness of causal relation probable to be a part of an answer by exerting term matching [H Fang; 2006], partial tree matching, and excitation polarity matching [C Hashimoto; 2014, 2012]. Term Matching approach matches the content words (nouns, verbs and adjectives) present in the question and the effect part of the answer to find appropriate causal relation to become an answer candidate. Partial tree matching approach finds causal relation to be an appropriate answer candidate by seeking whether its effect part comprises partial tree as in the question. Partial tree consists of atleast one content word. Excitation polarity matching perceives appropriate causal relation if its effect part contains atleast one noun having similar polarity as in the question.

The authors claimed that their approach endorsed causal relations with 80% precision, and outperformed in both precision and recall rates.

The system can be improved by matching inhibitory polarity (which suppress the effect and role of an entity).

Jong Hoon Oh 2016 [Oh; 2016] automatically generated training data of QA pairs using semi-supervised learning approach where causal relations are utilized. They employed supervised classifier to rank the answer passages, which were trained on the features i.e. morpho-syntactic, sentiment polarity, causal relation, excitation polarity features, and semantic word classes. The answers including causal relations are paraphrased that help to learn a wide range of causality expression patterns and recognize such causality expressions as candidates for answers to why-questions. This further improved 8% precision at the top answer over the QAS proposed by the authors in 2013 [Oh; 2013]. The system can further be enhanced by expanding the list of cue phrases, incorporating event causality [C Hashimoto; 2015], entailment/ causality patterns and zero anaphora resolution [R Mitkov 2014].

5.5.7. Metzler-Kanungo features:

Metzler and Kanungo in 2008 [D Metzler; 2008] provided six features to train answer re-ranker module. (1) ExactMatch assigns value 1 if query string is lexically matched with answer candidate, otherwise 0. (2) TermOverlap enumerates the count of query terms in a sentence by eliminating stop words. (3) SynonymsOverlap evaluates the number of original query terms and their synonyms in a sentence which is found by consulting WordNet. (4) LanguageModel computes the log likelihood of query terms, smoothed by Drichlet smoothing. (5) SentenceLength calculates the number of query terms in a sentence after removing stopwords. (6) SentenceLocation computes the position of sentence relative to the sentences in a document.

The authors have utilized semantic features with external resources like Wikipedia, WordNet, Word2vec semantic models to retrieve the most accurate answers which are lexically similar to a question. The work can be extended by incorporating new semantic features expressed at phrase or sentence level which are used to link two entities.

5.5.8. CNN & CTK method:

Kateryna Tymoshenko et. al. in 2016 [K. Tymoshenko; 2016] combines Convolution Tree Kernels (CTKs) [M Collins; 2002] and Convolutional Neural Networks (CNNs) [url] for ranking answer sentences in QAS. They investigated CTKs for learning classification and ranking functions. They proposed a unique deep learning approach for creating Question and answer passage pairs. Two sentence models were implemented on the basis of CNNs, which calculate semantic similarity score between two vectors of question ($x_q$) and document ($x_d$) given by:

$$sim(x_q, x_d) = x_q^T M x_d$$

where M is a similarity score.

$x'_d = M x_d$ is the mutation of candidate document, nearest to an input query $x_q$.

M is the similarity matrix.

To capture relational information, overlapping words are inserted into the word embedding and the enhanced words are passed through the layers to garble the correspondence between question and answer passage.

The authors implemented CTK and CNN in non-factoid QA, particularly on WikiQA and TREC13 which outperformed all other approaches. Also, CTKs are more efficient than CNNs, e.g., on TREC13, CTKs achieves MRR of 85.53 and MAP of 75.18 while CNNs achieves MRR of 77.93 and MAP of 71.09 as CTKs does syntactic parsing using two classifiers for question classification and named entity recognition whereas CNNs applies terms on unsupervised dataset.

5.5.9. Textual Entailment:

Harabagiu and Hickl in 2006 [S Harabagiu; 2006] incorporated textual entailment (TE) [DZ Korman; 2018] as a semantic relation to improve the accuracy of general domain QAS. When the question and their answer neither contain similar words nor their synonyms, then the appropriate answer is deduced by finding textual entailment between text segments. It is a semantic relation which derives one text fragment from the truth of other text fragment. Such entailing and entailed texts are termed text and hypothesis respectively. The authors presented three approaches to introduce textual entailment into QAS. (1) Finding entailment between question and candidate answers by removing those answers that don't meet minimum requirements based on entailment confidence ranging from 0 to 1. (2) Retrieved passages were ranked using entailment and entailment confidence was calculated between question and re-ranked passages. Final ranking of candidate answers were derived from features combining entailment confidence with keyword and relation-based features. (3) Questions submitted to QAS were entailed by Automatic Generated Questions which is used to extract answer passage associated with top-ranked entailed question.

When textual entailment was utilized to either refine or rank answer candidates, accuracy was improved from 32 to 52% in case answer type was detected whereas improvement from 30 to 40% in case no answer type was detected.

The results successfully show that textual entailment shall be incorporated into open domain Why-QAS to design operational answer validation system. Also, the effectiveness of QAS employing entailment patterns can be enhanced by using contradiction patterns to re-rank answers and also recognizing transitivity paths to obtain best combinations of QA pairs.

### 5.5.10. Lexical Semantic Language Model:

Lexical semantic model (Fried et. al.; 2015) is used to bridge the lexical chasm problem. Two types of lexical semantics are applied to QA, one is monolingual alignment model which learns associations between words appearing in QA pairs (Surdeanu et al; 2011; Yao et. al; 2013) and other computes semantic similarity between question and answers using language models (Yih et. al; 2013; Jansen et. al; 2014). The authors have used a large dataset of questions to evaluate both first-order and higher-order Lexical Semantic models. The re-ranking component analyses answer candidates to take-out both first and higher-order semantic features to be used in tandem with an environment to re-rank probable right answer candidates, to top positions.

The conclusion that can be drawn from this study was that all first order lexical models were able to outshine random and CR baselines (if can be explained). First order models could also outperform Jansen et. al. system [P Jansen; 2013]. Higher order models were able to perform well up to an order of 3 or 4. Under the scope of overall work on corpus of questions selected, higher order models were able to outperform the first order models.

| Literature | Methodology and Module | Techniques/Models | Comments |
|---|---|---|---|
| Verberne 2006 | Answer Re-ranking (BOW Model) | Bag of Words | Calculates overlap between Bag of Question items and Bag of Answer items |
| Jong-Hoon Oh 2013 | Answer Re-ranking (Morphosyntactic Analysis) | Morphological analyzer and syntactic dependency | Re-rank answer candidates using grammatical structure and ordering of words. Performance improves upon addition of semantic features |
| Jong-Hoon Oh 2013 | Answer Re-ranking (Semantic Word Class) | Semantic Word Classes: Collection of words in semantically similar context | Correspondence between word n-grams used in answer candidate and semantic word-class used in question to find relation between them, thus helping find the appropriate answer |
| Jong-Hoon Oh 2012 | Answer Re-ranking (Sentiment Analysis) | Identify sentiment polarities of words in answer candidates | Useful for opinionated QA, as answers to be extracted reflect opinion of words in question |
| Higashinaka and Isozaki 2008 | Answer Re-ranking (Content Similarity) | Based on probability of similarity in contents | Features address the meaning of the content involved in both question and answer candidate |
| Higashinaka and Isozaki 2008 | Answer Re-ranking (Causal Relation Features) | Train answer re-ranker candidates with: Causal relations, Causal expression and Content similarity | Consulted EDR dictionary to extract causal relations |
| Jong-Hoon Oh 2013 | Answer Re-ranking (Causal Relation Features) | Intra and Inter-sentential causal relations | This approach endorsed causal relations with high precission and outperformance in both precision and re-call rates |
| D Metzler 2008 | Answer Re-ranking | Six features to train re-ranker: Exact Match, TermOverlap, SynonymsOverlap, LanguageModel, SentenceLength, SentenceLocation | Use of semantic features with external resource to retrieve most accurate answers |
| K Tymoshenko 2016 | Answer Re-ranking | Convolution Neural Networks(CNN) and Convolution Tree Kernels(CTK) for re-ranking. Utilising deep learning for QA pairs | Implemented CTK and CNN on WikiQA and TREC13 with outperformance |
| Harabagiu | Answer Re- | Approach when | Accuracy was improved when |
| 2006 | ranking (Textual Entailment) | questions and answers do not contain similar words or synonyms | entailment was used to refine answer candidates. Effectiveness of QAS using entailment can be enhanced using contradiction patterns |
| Fred et. Al 2015 | Answer Re-ranking (Lexical Semantic Model) | Bridge the lexical chasm problem | Analyse answer candidates to derive both first and higher-order semantic features to re-rank right answer candidates |

*Table 10: Overview of the existing research on Answer Re-ranking module.*

| Challenges | Solutions |
|---|---|
| Choosing the appropriate features for finding most appropriate answer | Analysing the effect of different features on accuracy of QAS. Consider those features that play important role according to the scenario and improve the accuracy of the system |
| Assigning weights to the features | Sorting the features according to their impact on answering and assigning weights accordingly which can be done by training the neural network |
| Usage of different lexical words in question and best answer candidate. Sometimes best answer candidate doesn't use similar words as in a question and thus ranked lower than other answer candidates | Need of considering semantic features and assigning them more weightage than other features to return the highest ranked answer |

*Table 11: Challenges of Answer Re-ranker and their solutions*

## 6. Performance Metrics

The question answering systems (QASs) are evaluated by determining the correctness of answers. An answer is said to be correct if it belongs to the appropriate document and reacts to the question asked. There are some common measures which are used to gauge the performance of the system discussed below [O Kolomiyets; 2011] out of which MRR, Precision, Recall, F-measure are commonly used by researchers to compare their research with other previous researches.

### 6.1. Mean Reciprocal rank

The Reciprocal Rank (RR) is evaluated for an individual question which is measured by finding the reciprocal of the rank of first appropriate answer to a question. The score is 0 if no correct answer is found. If the relevant answer is found at rank 1, RR is calculated as 1, and 0.5 if a relevant answer is ranked second and so on.

Thus it is given as:
$$RR_{(q_i)} = 1/r_i$$
$$RR(q_i) = 1 / \{\text{rank of first correct answer for } q_i\}$$
MRR is mean reciprocal rank which computes the capability of the system to retrieve answers for the set of N questions by taking the mean of the reciprocal ranks and is calculated as:
$$MRR = \Sigma^n_{i=1} RR_i / n \text{ (where n is the number of questions)}$$
Thus it depicts how early the relevant results are obtained in ranking. Its considered perfect if its value is close to 1 and worse if close to 0. Thus more the MRR, more the accuracy of the system.

### 6.2. Precision
Precision is one of the traditional measures used in Information Retrieval which determines the number of retrieved relevant documents. It is defined as the fraction of answer documents retrieved which are relevant to the user's question. Its formula is given as:

Precision = no. of relevant answer documents retrieved / no. of retrieved answer documents

= P( relevant answers | retrieved answers)

Good precision is just an indication of good accuracy. There may be the possibility of random and systematic errors from the system. In the case of random errors, good precision means good accuracy but the presence of systematic errors prevents us from concluding that good precision denotes good accuracy.

### 6.3. Precision at n (P@n)
Precision@n finds out relevant top-n answer documents. These top-n are the first n ranked answers corresponding to the question. Here, n corresponds to a number of documents shown to the user that is, Precision@1 means only the first document is manifested to user, Precision@5 extracts first 5 results to users. It is given as:

Precision@n = no. of recommended answers @n that are relevant / no. of recommended answers @n)

This alone doesn't play major role in measuring accuracy of the system but rather it is utilized in other metric which is Mean Average Precision.

### 6.4. MAP (Mean Average Precision)
It calculates the mean of the average precision scores for each question and thus helps to determine the overall quality of the top-n answer candidates to a question.

$$MAP = \frac{1}{|Q|} \sum_{q \in Q} \frac{\sum_{k=1}^{n}(\Pr ec(k) \times rel(k))}{|A_q|}$$

Q – set of questions,
$A_q$ – set of correct answers to a question, q ∈ Q
Prec(k) – precision at cut-off k in the top-n answer candidates

Rel(k) – indicator, which is rated as 1 if the item at rank k is a correct answer in $A_q$

Thus, it is an effective indicator than Precision as performance is measured on a whole for a system which finds the precision scores for each question asked in QAS, rather than finding for a single question.

### 6.5. Recall

Recall also termed as sensitivity which evaluates the fraction of answer documents relevant to the question that are auspiciously retrieved. It is stated as the probability of relevant document retrieved and is calculated by dividing the number of relevant documents retrieved by the total relevant documents.

Recall = |{relevant documents} ∩ {retrieved documents}| / |{relevant documents}|.

Thus, it measures the ability to find all relevant results to a query. We tried to increase its value to improve the accuracy.

### 6.6. Recall @k:

It is similar to recall but calculates the proportion of relevant documents that are ranked in the top-K answer documents corresponding to a question.

### 6.7. F-score

Since, it is seen that the value of both precision and recall is increased to improve accuracy, thus there comes an issue while assigning weights to precision and recall i.e. which metric is given more weight than other, thus to maintain the balance between both precision and recall, F-score or F-measure is calculated which measures the harmonic mean of precision and recall given as:

F = (2* precision * recall) / (precision+recall)

$F_1$ measure uniformly weighs recall and precision, $F_2$ measure weights recall twice as much as precision, and similarly $F_{0.5}$ measure weights precision twice as much as recall which is generalized for a non-negative real ß values as:

$F_ß$ = (1+ß$^2$) * (precision*recall) / (ß$^2$ * precision +recall)

Thus, depending on the type of errors to be minimized, $F_2$ and $F_{0.5}$

## Conclusion

Through the literature survey we have enlisted the existing research on Why-type Question Answering System. The improvement in accuracy of one module can lead to an overall improved performance for the complete system. The survey describes the work undertaken and techniques used on various modules, the accuracy of such techniques and future enhancements to improve accuracy of these modules.

The arguments common to general QAS's viz. understanding of natural language questions and processing them accurately for relevant correct answers remain an important concern for Why-QAS as well, so does the selection of relevant documents from which the answer is looked into.

An important research area/challenge would be to understand the focus and meaning (semantics) of the question. As there exists a vocabulary gap between the words used in question and its corresponding answer, the need to better understand the semantics of question and it's processing to determine its correct meaningful answer.

Another challenge is encountered in re-ranking answer candidates. Score is assigned to the candidate set of answers based on set of features on which classifier is trained. There are various features like morpho-syntactic, bag-of-words, causal relations, sentiment polarities and word classes which play an important role in determining appropriate answer.

With each module of QAS, different challenges and their solutions have been discussed which will aid the research community to work on the issues and try to improve the performance of the system.